\pdfoutput=1

\documentclass[11pt]{article}

\usepackage{emnlp2022}

\usepackage{times}
\usepackage{latexsym}

\usepackage[T1]{fontenc}

\usepackage[utf8]{inputenc}

\usepackage{microtype}

\usepackage{inconsolata}

\usepackage{arydshln}
\usepackage{xcolor}
\usepackage{booktabs}
\usepackage{multicol}
\usepackage{multirow}
\usepackage{graphicx}
\usepackage{tabularx}
\usepackage{float}
\usepackage{bm}
\usepackage{todonotes}
\usepackage{xargs}
\usepackage{xspace}
\usepackage{amsmath}
\usepackage{cleveref}
\usepackage{subcaption}
\usepackage{mathtools}
\usepackage{bigstrut}
\usepackage{tikz}
\usetikzlibrary{shapes}
\usepackage{rotating}
\usepackage{siunitx}
\usepackage{dcolumn}
\newcolumntype{L}{D{.}{.}{2,1}}
\newcolumntype{P}{D{\pm}{\pm}{2,1}}

\renewcommand*{\backrefalt}[4]{%
    \ifcase #1 \footnotesize{(Not cited.)}%
    \or        \footnotesize{(Cited on page~#2)}%
    \else      \footnotesize{(Cited on pages~#2)}%
    \fi}

\newcommand*{\eg}{e.g.\@\xspace}
\newcommand*{\ie}{i.e.\@\xspace}

\makeatletter\newcommand*{\etc}{%
	\@ifnextchar{.}%
	{etc}%
	{etc.\@\xspace}%
}
\makeatother

\newcommand{\glat}{\textsc{GLAT}\@\xspace}
\newcommand{\ctc}{\textsc{CTC}\@\xspace}

\newcommand{\ds}{\textsc{DS}\@\xspace}
\newcommand{\bert}{\textsc{BERT}\@\xspace}
\newcommand{\adam}{\textsc{Adam}\@\xspace}

\newcommand{\lenglish}{\textsc{EN}}
\newcommand{\lromanian}{\textsc{RO}}
\newcommand{\lgerman}{\textsc{DE}}

\newcommand{\lende}{\lenglish$\rightarrow$\lgerman\@\xspace}
\newcommand{\lenro}{\lenglish$\rightarrow$\lromanian\@\xspace}
\newcommand{\ldeen}{\lgerman$\rightarrow$\lenglish\@\xspace}
\newcommand{\lroen}{\lromanian$\rightarrow$\lenglish\@\xspace}
\newcommand{\lendeb}{\lenglish$\leftrightarrow$\lgerman\@\xspace}
\newcommand{\lenrob}{\lenglish$\leftrightarrow$\lromanian\@\xspace}

\DeclareMathOperator*{\argmax}{arg\,max}

\definecolor{red}{RGB}{141,45,57}
\definecolor{dark}{RGB}{55,65,74}
\definecolor{blue}{RGB}{0,105,170}
\definecolor{gold}{RGB}{174,159,109}
\definecolor{gray}{RGB}{175,179,183}
\definecolor{darkgreen}{RGB}{50,110,30}

\makeatletter
\newcommand*\standardbin{+}
\newcommand*\tabularbin[1]{%
  \mathbin{\mathpalette{\@tabularsym\standardbin}{#1}}%
}
\newcommand*\@tabularsym[3]{%
  \setbox\z@\hbox{$#2#1\m@th$}%
  \hbox to\wd\z@{\hss$#2#3\m@th$\hss}%
}
\makeatother

\newcommand{\chd}[1]{(\textcolor{red}{$\mathbf{\tabularbin-#1}$})} 
\newcommand{\chu}[1]{(\textcolor{darkgreen}{$\mathbf{\tabularbin+#1}$})} 
\newcommand{\che}[1]{(\textcolor{gold}{$\mathbf{\tabularbin+#1}$})} 

\newcommand{\sacrebleu}{\texttt{sacreBLEU}\xspace}
\newcommand{\mtcompare}{\texttt{compare-mt}\xspace}
\newcommand{\bleu}{\textsc{Bleu}\xspace}
\newcommand{\chrf}{\textsc{chrF++}\xspace}
\newcommand{\ter}{\textsc{Ter}\xspace}
\newcommand{\comet}{\textsc{Comet}\xspace}
\newcommand{\fairseq}{\texttt{fairseq}\xspace}

\newcommand{\insig}{\emph{non-significant}\xspace}

\newcommandx{\robin}[2][1=]{\vspace{0.2cm} \todo[linecolor=blue,backgroundcolor=blue!15,bordercolor=blue, #1]{\textbf{Robin:} #2}}
\newcommandx{\telmo}[2][1=]{\vspace{0.2cm} \todo[linecolor=gold,backgroundcolor=gold!25,bordercolor=gold, #1]{\textbf{Telmo:} #2}}
\newcommandx{\stephan}[2][1=]{\vspace{0.2cm} \todo[linecolor=darkgreen,backgroundcolor=darkgreen!25,bordercolor=darkgreen, #1]{\textbf{Stephan:} #2}}

\newcommand{\todotemplate}{\textcolor{red}{TODO}\xspace}
\newcommandx{\trobin}[2][1=]{\vspace{0.2cm} \todo[linecolor=blue,backgroundcolor=blue!15,bordercolor=blue, #1]{\textbf{\todotemplate @Robin:} #2}}
\newcommandx{\ttelmo}[2][1=]{\vspace{0.2cm} \todo[linecolor=gold,backgroundcolor=gold!25,bordercolor=gold, #1]{\textbf{\todotemplate @Telmo:} #2}}
\newcommandx{\tstephan}[2][1=]{\vspace{0.2cm} \todo[linecolor=darkgreen,backgroundcolor=darkgreen!25,bordercolor=darkgreen, #1]{\textbf{\todotemplate @Stephan:} #2}}

\makeatletter
\newcommand{\ssymbol}[1]{^{\@fnsymbol{#1}}}
\newcommand{\nosig}{\ssymbol{2}\xspace}
\makeatother

\makeatletter
\def\adl@drawiv#1#2#3{%
        \hskip.5\tabcolsep
        \xleaders#3{#2.5\@tempdimb #1{1}#2.5\@tempdimb}%
                #2\z@ plus1fil minus1fil\relax
        \hskip.5\tabcolsep}
\newcommand{\cdashlinelr}[1]{%
  \noalign{\vskip\aboverulesep
           \global\let\@dashdrawstore\adl@draw
           \global\let\adl@draw\adl@drawiv}
  \cdashline{#1}
  \noalign{\global\let\adl@draw\@dashdrawstore
           \vskip\belowrulesep}}
\makeatother

\title{Non-Autoregressive Neural Machine Translation: A Call for Clarity}

\author{Robin M.~Schmidt \quad Telmo Pessoa Pires \quad Stephan Peitz \quad Jonas Lööf \\
  Apple\\
  \texttt{\{robin\_schmidt, telmo, speitz, jloof\}@apple.com}}

\begin{document}
\maketitle
\begin{abstract}
Non-autoregressive approaches aim to improve the inference speed of translation models by only requiring a single forward pass to generate the output sequence instead of iteratively producing each predicted token. Consequently, their translation quality still tends to be inferior to their autoregressive counterparts due to several issues involving output token interdependence. In this work, we take a step back and revisit several techniques that have been proposed for improving non-autoregressive translation models and compare their combined translation quality and speed implications under third-party testing environments. We provide novel insights for establishing strong baselines using length prediction or CTC-based architecture variants and contribute standardized \bleu, \chrf, and \ter scores using \sacrebleu on four translation tasks, which crucially have been missing as inconsistencies in the use of tokenized \bleu lead to deviations of up to 1.7 \bleu points. Our open-sourced code is integrated into \fairseq for reproducibility.\footnote{\url{https://github.com/facebookresearch/fairseq/pull/4431}}
\end{abstract}

\section{Introduction}
Traditional sequence-to-sequence models aim to predict a target sequence $e_1^I=e_1, \ldots, e_i, \ldots, e_{I}$ of length $I$ given an input sequence $f_1^J=f_1, \ldots, f_j, \ldots, f_{J}$ of length $J$. In the autoregressive case, this is done token by token, and the probability distribution for the output at timestep $i$ is conditioned on the source sentence $f_1^J$ but also on the preceding outputs of the model $e_1^{i-1}$, and parameterized by $\bm{\theta}$:

\begin{equation}
    p_{\bm{\theta}}(e_1^I | f_1^J) = \prod_{i=1}^I p_{\bm{\theta}}(e_i|e_1^{i-1}, f_1^J).
    \label{eq.at}
\end{equation}
Even though these types of models are widely deployed, one of the major drawbacks is the inherent left-to-right factorization that requires iterative generation of output tokens, which is not efficiently parallelizable on modern hardware such as GPUs or TPUs. Non-autoregressive translation, on the other hand, assumes conditional independence between output tokens, allowing all tokens to be generated in parallel. Effectively, it removes the dependence on the decoding history for generation: 
\begin{equation}
    p_{\bm{\theta}}(e_1^I | f_1^J) = \prod_{i=1}^I p_{\bm{\theta}}(e_i|f_1^J).
    \label{eq.nat}
\end{equation}
This modeling assumption, despite its computational advantages, tends to be more restrictive and introduces the \emph{multimodality problem} \citep{Gu0XLS18} where consecutive output tokens are repeated or fail to correctly incorporate the preceeding information to form a meaningful translation. Overcoming this problem is one of the key challenges to achieving parity with autoregressive translation on a wide variety of tasks and architectures.

Closest to the presented work is a study by \citet{gu-kong-2021-fully} that also analyzes recent works regarding their effectiveness and successfully combines them to achieve remarkable translation quality. In this work, we additionally address several shortcomings in the literature: 1) standardized \bleu, \chrf, and \ter scores using \sacrebleu accompanied by open source \fairseq code; 2) more realistic non-autoregressive speed-up expectations, by comparing against faster baselines; 3) a call for clarity in the community regarding data pre-processing and evaluation settings.

\section{Experimental setup}
\label{sec.datasets}
\paragraph{Datasets \& knowledge distillation}
We perform our experiments on two datasets: WMT'14 English$\leftrightarrow$German (\lenglish $\leftrightarrow$\lgerman{}, 4M sentence pairs), and WMT'16 English$\leftrightarrow$Romanian (\lenglish $\leftrightarrow$\lromanian{}, 610K sentence pairs), allowing us to evaluate on $4$ translation directions. As is common practice in the non-autoregressive literature, we train our models with sequence-level knowledge distillation (KD, \citealp{kim-rush-2016-sequence}), obtained from Transformer \emph{base} teacher models \citep{VaswaniSPUJGKP17} with a beam size of $5$. We tokenize all data using the Moses tokenizer and also apply the Moses scripts \citep{koehn-etal-2007-moses} for punctuation normalization. Byte-pair encoding (BPE, \citealp{sennrich-etal-2016-neural}) is used with $40,000$ merge operations.

For \lenrob, we use WMT'16 provided scripts to normalize the \lromanian{} side, and to remove diacritics for \lroen only. \lenro keeps diacritics for producing accurate translations which explains the gap between our numbers and those claimed in previous works \citep{gu-kong-2021-fully,qian-etal-2021-glancing}, who besides computing \bleu on tokenized text, compared on diacritic-free text. More details are summarized in \Cref{app:dataset_details}.

\paragraph{Models}
In this work, we focused our efforts on four models/techniques for training non-autoregressive transformer (NAT) models: Vanilla NAT (\citealp{Gu0XLS18}), Glancing Transformer (GLAT, \citealp{qian-etal-2021-glancing}), Connectionist Temporal Classification (\ctc, \citealp{GravesFGS06,libovicky-helcl-2018-end}), and Deep Supervision (DS, \citealp{huang2021nonautoregressive}), see \Cref{app:related_work,app:methods} for a short overview. For all NAT models, we use learned encoder and decoder positional embeddings, shared word embeddings \citep{press-wolf-2017-using}, no label-smoothing, keep \adam betas and epsilon as defaults with $(0.9, 0.98)$ and $1\mathrm{e}{-8}$ respectively, train for $200$k (\lendeb) / $30$k (\lenrob) update steps with $10$k / $3$k warmup steps using an inverse square root schedule \citep{VaswaniSPUJGKP17} where the $5$ best checkpoints are averaged based on validation \bleu. For \ctc-based models, the source upsampling factor is $2$. More details can be found in our open-sourced training procedures. We intentionally did not fine-tune hyperparameters for each translation direction and instead opted for choices that will most likely transfer across datasets.

For autoregressive transformer (AT) models, we focus on the \emph{base} \citep{VaswaniSPUJGKP17} architecture, as well as a deep encoder ($11$ layers) \& shallow decoder ($2$ layers) variant \citep{Kasai0PCS21}.

\section{A call for clarity}
From our experiments with different algorithms, we noticed a few problems that are currently not properly addressed in the literature. These issues make it harder to do a rigorous comparison, so here, we explain how we addressed them, and give recommendations for future work.

\begin{table*}[t]
\setlength\tabcolsep{4pt}
\centering
\scriptsize
\begin{tabular}{llLLLLLrS[table-format=3.1]@{${}\pm{}$}S[table-format=1.2]rS[table-format=4.1]@{${}\pm{}$}S[table-format=2.1]}
\toprule
 \multicolumn{1}{l}{\multirow{2}{*}{\textbf{Models}}} & \multicolumn{1}{c}{\multirow{2}{*}{$|\bm{\theta}|$}} & 
 \multicolumn{1}{c}{\textbf{WMT'13}} & \multicolumn{2}{c}{\textbf{WMT'14}} & \multicolumn{2}{c}{\textbf{WMT'16}} &
 \multicolumn{1}{c}{\multirow{1}{*}{\textbf{Speed}}} & \multicolumn{2}{c}{\multirow{1}{*}{\textbf{Latency GPU}}} & \multicolumn{1}{c}{\multirow{1}{*}{\textbf{Speed}}} & \multicolumn{2}{c}{\multirow{1}{*}{\textbf{Latency CPU}}} \\
 \multicolumn{1}{c}{} & & \multicolumn{1}{c}{\textbf{\lende}} & \multicolumn{1}{c}{\textbf{\lende}} & \multicolumn{1}{c}{\textbf{\ldeen}} & \multicolumn{1}{c}{\textbf{\lenro}} & \multicolumn{1}{c}{\textbf{\lroen}} & \multicolumn{1}{c}{(GPU)} & \multicolumn{2}{c}{(\si{\ms}~/~sentence)} &
 \multicolumn{1}{c}{(CPU)} & 
 \multicolumn{2}{c}{(\si{\ms}~/~sentence)}\\
\midrule
\textbf{Autoregressive} & & & & & & & & \multicolumn{1}{c}{} & & & \multicolumn{1}{c}{} & \\
\midrule

 Transformer \textit{base} & 66M & 25.6 & 26.2 & 30.3 & 23.3 & 32.8 & 0.5$\times$ & 165.8 & 1.3  & 0.3$\times$ & 657.2 & 11.4 \\
 \quad\quad + Beam $5$ (teacher) & 66M & 26.4 & 26.8 & 31.5 & 24.0 & 33.4 & 0.4$\times$ & 212.8 & 8.9  & 0.1$\times$ & 1272.4 & 38.5 \\
 \quad\quad + KD & 66M & 26.2 & 26.5 & 30.6 & 23.8 & 32.6\nosig & 0.5$\times$ & 173.4 & 1.0  & 0.3$\times$ & 646.3 & 8.9 \\
\cdashlinelr{1-13}
 Transformer \textit{base} (11-2) & 63M $\downarrow$ & 25.6\nosig & 26.1\nosig & 30.1\nosig & 23.8 & 32.7\nosig & 0.9$\times$ & 89.1 & 3.2  & 0.4$\times$ & 424.2 & 15.0 \\
 \quad\quad + KD & 63M $\downarrow$ & 26.3 & 26.7 & 30.8 & 23.8\nosig & 32.7\nosig & 0.9$\times$ & 88.1 & 1.5  & 0.5$\times$ & 408.0 & 3.2 \\
 \quad\quad + KD + AA & 62M $\downarrow$ & 26.2\nosig & 26.7\nosig & 30.7\nosig & 23.9\nosig & 32.6\nosig & 1.0$\times$ & 79.4 & 2.2  & 0.5$\times$ & 396.9 & 5.4 \\
 \quad\quad + KD + AA + SL & 62M $\downarrow$ & 26.2\nosig & 26.7\nosig & 30.7 & 23.9\nosig & 32.6 & 1.0$\times$ & 82.7 & 7.5  & 1.0$\times$ & 188.3 & 2.2 \\

\midrule
\textbf{Non-Autoregressive} & & & & & & & & \multicolumn{1}{c}{} & & & \multicolumn{1}{c}{} & \\
\midrule

 Vanilla-NAT & 66M & 21.8 & 21.2 & 26.7 & 19.3 & 27.4 & 6.1$\times$ & 13.1 & 0.3  & 2.1$\times$ & 90.1 & 0.4 \\
 \quad\quad + \glat & 66M & 24.2 & 24.1 & 28.9 & 21.3 & 29.4 & 6.2$\times$ & 12.8 & 0.4  & 2.1$\times$ & 90.4 & 0.6 \\
\cdashlinelr{1-13}
 \ctc & 66M & 25.1 & 25.5 & 29.8 & 23.0 & 32.3 & 6.4$\times$ & 12.4 & 0.4  & 1.5$\times$ & 122.9 & 1.7 \\
 \quad\quad + \ds & 66M & 25.3\nosig & 25.6\nosig & 30.0\nosig & 22.8\nosig & 32.7 & 6.3$\times$ & 12.7 & 0.6  & 1.6$\times$ & 121.4 & 0.5 \\
 \quad\quad + \glat  & 66M & 25.8 & 25.7 & 30.4 & 23.2\nosig & 32.8 & 6.3$\times$ & 12.7 & 0.3  & 1.6$\times$ & 121.2 & 1.2 \\
 \quad\quad + \glat + \ds & 66M & 25.9\nosig & 25.9\nosig & 30.7 & 23.2\nosig & 32.7\nosig & 6.2$\times$ & 12.8 & 0.3  & 1.5$\times$ & 121.8 & 0.4 \\
\cdashlinelr{1-13}
 \ctc (11-2) & 63M $\downarrow$ & 25.4 & 25.6 & 30.1 & 23.3 & 32.8 & 6.9$\times$ & 11.5 & 0.2  & 1.6$\times$ & 114.6 & 1.3 \\
\quad\quad + \glat & 63M $\downarrow$ & 25.7 & 25.8\nosig & 30.7 & 23.5\nosig & 33.1\nosig & 6.8$\times$ & 11.7 & 0.4  & 1.6$\times$ & 114.6 & 1.2 \\
 \quad\quad + \glat + SL & 63M $\downarrow$ & 25.7\nosig & 25.8\nosig & 30.7 & 23.5\nosig & 33.1\nosig & 6.7$\times$ & 11.9 & 0.2  & 2.6$\times$ & 72.5 & 0.8 \\

\bottomrule
\end{tabular}
\caption{Main results using \sacrebleu on four different translation directions. Latencies are measured on the WMT'14 \lende test set as the average processing time per sentence using batch size $1$. All AT models require $I$ decoding iterations at inference time while the NAT models only require $1$. WMT'13 \lende serves as validation performance. No re-scoring is used for all results. Statistically \insig results using paired bootstrap resampling are marked with $\nosig$ for $p \geq 0.05$, see \Cref{sec:stat_tests} for more details. \lenro keeps diacritics and $\downarrow$ indicates less number of parameters than the baseline.}
\label{tb.results_full_bleu}
\end{table*}


\subsection{Evaluation: The return of \sacrebleu}
\label{sec.broken}
In many works, the method for comparing novel non-autoregressive translation models is \emph{tokenized} \bleu \citep{papineni-etal-2002-bleu}. However, this can cause incomparable translation quality scores as the reference processing is \emph{user-supplied} and not \emph{metric-internal}. Well known problems with user-supplied processing for \bleu include different tokenization and normalization schemes applied to the reference across papers that can cause \bleu deviations of up to 1.8 \bleu points \citep{post-2018-call}. For the NAT literature, in particular, the basis for comparison has been the distilled data released by \citet{ZhouGN20} where binarization\footnote{This refers to converting the training data into binary format \ie producing the output of \texttt{fairseq-preprocess}.} is handled by the individual researchers. One instance for such deviations can be found for the \bleu computations on WMT'14 \lende of \citet{huang2021nonautoregressive} using processed references instead of the original ones. Applying a vocabulary obtained from the distilled training data for processing the references results in \texttt{<unk>} tokens appearing in the references for the German left (\,\emph{„}\xspace) and right (\emph{“}\,) quotation marks because these tokens have not been generated during the distillation process. However, using such processed references leads to artificially inflated \bleu scores. We evaluated our models and open-sourced models of previous works using processed references, original references, and \sacrebleu and observe deviations of up to 1.7 \bleu points across multiple previous works. This fact, together with arbitrary optimization choices that can not be accredited to the presented method, hinders meaningful progress in the field by making results incomparable across papers without individual re-implementation by subsequent researchers. 

\paragraph{Our approach}
To alleviate these problems, we urge the community to return to \sacrebleu\footnote{\url{https://github.com/mjpost/sacrebleu}} \citep{post-2018-call} for evaluation. We provide standardized \bleu \citep{papineni-etal-2002-bleu}, \chrf \citep{popovic-2017-chrf}, and \ter \citep{snover-etal-2006-study} scores in \Cref{tb.results_full_bleu,tb.results_full_chrf,tb.results_full_ter} for current state-of-the-art non-autoregressive models that are reproducible with our codebase, as well as easy to configure through flags, and can be used as baselines for novel model comparisons. No re-ranking is used for all numbers. See \Cref{app:signatures} for the used \sacrebleu evaluation signatures. Statistically \insig results ($p \geq 0.05$) are marked with $\nosig$ for all main results using paired bootstrap resampling \citep{koehn-2004-statistical}. For more information on how we select base and reference systems for these statistical tests, please see \Cref{sec:stat_tests}. As such, we directly try to address the issues found by \citet{marie-etal-2021-scientific} for the non-autoregressive translation community. 

\subsection{Establishing a realistic AT speed baseline}
Traditionally, speed comparisons in the NAT literature compare to weak baselines in terms of achievable autoregressive decoding speed which leads to heavily overestimated speed multipliers for non-autoregressive models which has been recently criticized by \citet{heafield-etal-2021-findings}.

\paragraph{Our approach}
We deploy a much more competitive deep encoder ($11$ layers) \& shallow decoder ($2$ layers) \citep{Kasai0PCS21} autoregressive baseline, with average attention (AA, \citealp{zhang-etal-2018-accelerating}) and shortlists (SL, \citealp{schwenk-etal-2006-continuous,junczys-dowmunt-etal-2018-marian-cost}). It achieves a similar accuracy with comparable number of parameters in comparison to the \emph{base} architecture (see \Cref{tb.results_full_bleu}).

Our latency measurements (mean processing time per sentence over $3$ runs) were collected using a single NVIDIA A100 GPU (Latency GPU) or a single-threaded Intel Xeon Processor (Skylake, IBRS) @ 2.3 GHz (Latency CPU), both with batch size of $1$ which faithfully captures the inference on a deployed neural machine translation model.

\subsection{Establishing an accurate NAT baseline}
Hyperparameters and smaller implementation details are often not properly ablated and only copied between open sourced training instructions, causing ambiguity for researchers on which parts of the proposed approach \emph{actually} make a difference. Such choices include the decoder inputs and many more optimization configurations such as dropout values, initialization procedures, and activation functions. Here, we provide our insights from experimenting with these variants.

\subsubsection{No source embeddings as decoder input}
It is common in NAT models to feed the source embeddings as decoder inputs using either \emph{UniformCopy}, \emph{SoftCopy} \citep{wei-etal-2019-imitation}, or sometimes even more exotic approaches without any details such as an ``attention version using positions'' \citep{qian-etal-2021-glancing}, in this paper referenced as \emph{PositionAttention}. While it seems to be common consensus that using any of these variants benefits the final translation quality of the model, our experiments with these have shown the clear opposite when compared to simply feeding the embedding of a special token such as \texttt{<unk>} (see \Cref{tb.src_embed_results}). 

\begin{figure*}
\centering

\begin{subfigure}{0.48\textwidth}
    \includegraphics[width=\textwidth]{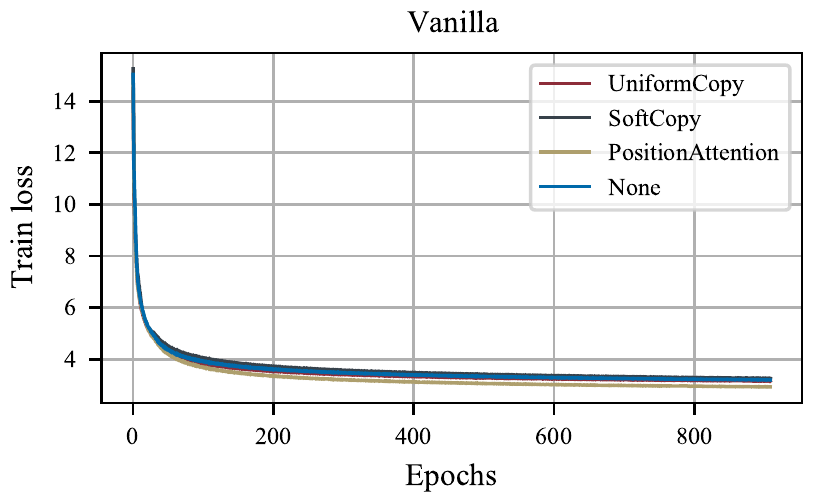}
\end{subfigure}
\hfill
\begin{subfigure}{0.48\textwidth}
    \includegraphics[width=\textwidth]{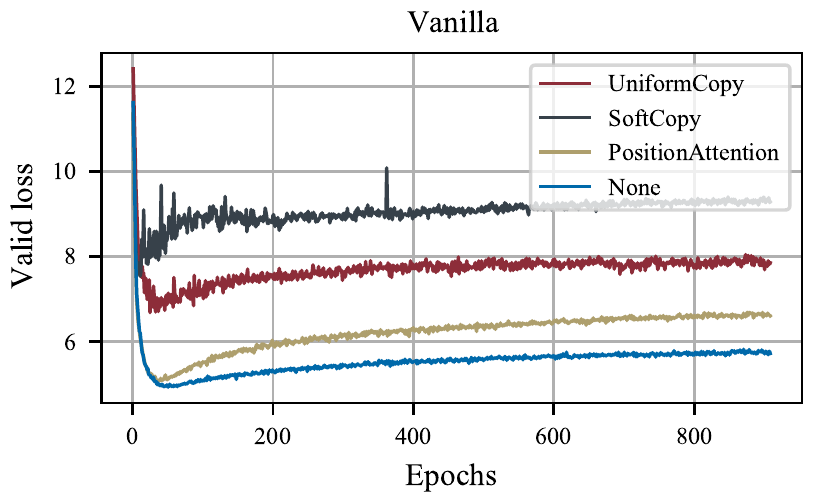}
\end{subfigure}

\begin{subfigure}{0.48\textwidth}
    \includegraphics[width=\textwidth]{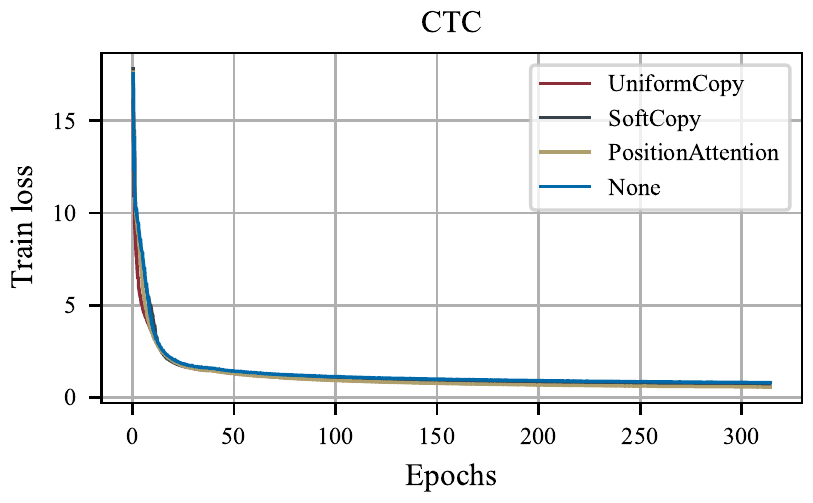}
\end{subfigure}
\hfill
\begin{subfigure}{0.48\textwidth}
    \includegraphics[width=\textwidth]{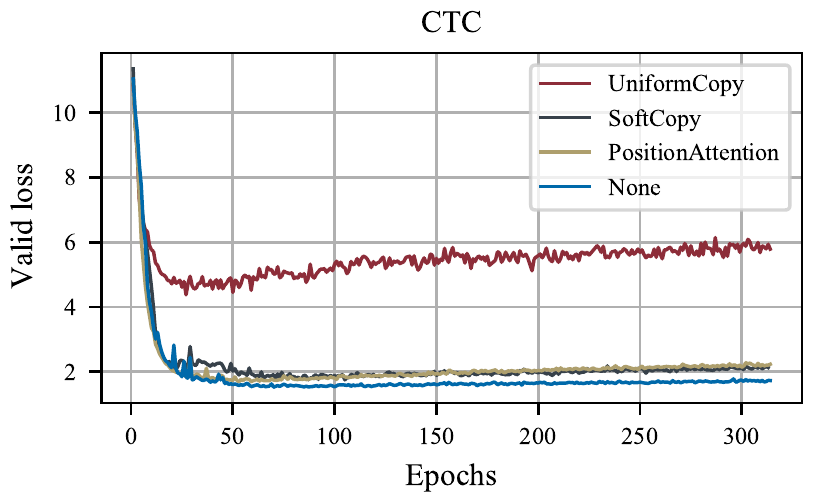}
\end{subfigure}
\caption{Influence of guided decoder input for VanillaNAT (top) and \ctc (bottom) on the training (left) and validation (right) loss using the WMT'13 \lende evaluation set.}
\label{fig.src_embed_curves}
\end{figure*}

In our experiments, we could not discern any consistent improvement on the final translation quality from any of the methods. For \emph{SoftCopy}, the learned temperature parameter has a strong influence on translation quality and is hard for the network to train. Notably, for Vanilla NAT and \glat the effects of adding any such decoder input is much more severe than for \ctc-based models. One of the only cases where we see improvement is when adding \emph{PositionAttention} to the Vanilla NAT model but since it does not show consistent \bleu gains for \glat and \ctc models, we do not investigate this any further. Looking at the validation losses from \Cref{fig.src_embed_curves}, one can observe higher fluctuations for any variant of guided decoder input compared to disabling it. 


\begin{table}[t]
\centering
\small
\tabcolsep 4pt
\begin{tabular}{lcc}
\toprule
\multirow{2}{*}{\textbf{Models}} & \textbf{WMT'13} & \textbf{WMT'14} \\
 & \textbf{\lende} & \textbf{\lende} \\
 \midrule
 \textbf{Vanilla NAT}  & \multirow{2}{*}{22.2} & \multirow{2}{*}{20.8} \\ 
 \citep{Gu0XLS18} & & \\[0.1cm]
 \quad\quad + UniformCopy & 20.4 \chd{1.8} & 19.9 \chd{0.9} \\ 
 \quad\quad + SoftCopy & 15.3 \chd{6.9} & 14.2 \chd{6.6} \\ 
 \quad\quad + PositionAttention & 22.3 \chu{0.1} & 21.5 \chu{0.7} \\ 
 \midrule
 \textbf{\glat} & \multirow{2}{*}{24.3}  & \multirow{2}{*}{23.9} \\ 
 \citep{qian-etal-2021-glancing} & & \\[0.1cm]
 \quad\quad + UniformCopy & 20.0 \chd{4.3} & 19.3 \chd{4.6} \\ 
 \quad\quad + SoftCopy & 16.6 \chd{7.7} & 15.1 \chd{8.8} \\ 
 \quad\quad + PositionAttention & 24.2 \chd{0.1} & 24.1 \chu{0.2} \\ 
 \addlinespace
 \midrule
 \textbf{\ctc} & \multirow{2}{*}{25.0} & \multirow{2}{*}{25.5} \\ 
 \citep{saharia-etal-2020-non} & & \\[0.1cm]
 \quad\quad + UniformCopy & 22.5 \chd{2.5} & 23.4 \chd{2.1} \\ 
 \quad\quad + SoftCopy & 24.9 \chd{0.1} & 25.5 \che{0.0} \\ 
 \quad\quad + PositionAttention & 24.8 \chd{0.2} & 25.0 \chd{0.5} \\ 
 \bottomrule
\end{tabular}
\caption{\bleu scores when using different source embedding copying strategies. Distilled training data and hyperparameters stay consistent.}
\label{tb.src_embed_results}
\end{table}

Interestingly, for a setup without guided decoder input, especially the token embedding as well as the self-attention component in the first decoder layer supposedly contain only little information since only consecutive special tokens \ie \texttt{<unk>}'s are passed. We investigate the performance contribution of these in \Cref{tb.decoder_components} and observe that 1) the self-attention component in the first decoder layer only has minor impact and 2) solely relying on learned positional embeddings is sufficient for achieving comparable performance. However, we do keep both components in the main results for comparability reasons.

\begin{table}[t]
\centering
\small
\tabcolsep 4pt
\begin{tabular}{lcc}
\toprule
\multirow{2}{*}{\textbf{Models}} & \textbf{WMT'13} & \textbf{WMT'14} \\
 & \textbf{\lende} & \textbf{\lende} \\
\midrule
 \textbf{\ctc} & \multirow{2}{*}{24.9} & \multirow{2}{*}{25.8} \\ 
 \citep{saharia-etal-2020-non} & & \\[0.1cm]
   \quad\quad + no layer 0 self attention & 25.0 \chu{0.1} & 25.5 \chd{0.3} \\ 
 \quad\quad + no decoder embeddings & 24.7 \chd{0.2} & 25.8 \chd{0.2} \\ %
 \quad\quad + both & 24.9 \che{0.0} & 25.5 \chd{0.3} \\ 
 \midrule
 \textbf{\ctc + \glat} & \multirow{2}{*}{25.8} & \multirow{2}{*}{26.0} \\  
 \citep{qian-etal-2021-glancing} & & \\[0.1cm] 
 \quad\quad + no layer 0 self attention & 25.8 \che{0.0} & 25.9 \chd{0.1} \\ %
 \bottomrule
\end{tabular}
\caption{Overview of the translation quality implications when removing decoder components.}
\label{tb.decoder_components}
\end{table}

\subsubsection{Influence of dropout}
Dropout is one of the hyperparameters which either prior works did not mention or simply used the default values from previous studies. We performed a sweep for different architectures and language pairs, where we varied the dropout probability from $0$ to $0.5$, with increments of $0.1$ (see \Cref{fig.drop_swp}). We observe that choosing an appropriate dropout rate is important for achieving high translation quality. In particular, for \lendeb, $0.1$ achieves the best validation performance for all models, confirming the choice in prior works. For \lenrob, we find that the value used in prior works, $0.3$, is optimal for non-\ctc models, but, for \ctc-based approaches, higher values seem to outperform.

\subsubsection{Miscellaneous optimization choices}
Many previous works use the Gaussian Error Linear Unit (GELU, \citealp{hendrycks2016gelu}) activation function, initialize the parameters with the procedure from \bert \citep{devlin-etal-2019-bert}, use length offset prediction, or a larger
\adam \citep{KingmaB14} $\epsilon$ value for optimization. In \Cref{app:adam_eps} we summarize the effect of these choices and find that for Vanilla NAT models these variations regularly have a positive impact on translation quality but for \ctc-based models the picture is less clear and rarely leads to consistent \bleu improvements. Based on these results, we deploy all choices besides $\epsilon$ tuning for non-\ctc variants but omit them for \ctc-based models.

\subsection{Model comparison}
\label{sec.results}

Our results are consistent across the language pairs and metrics shown in \Cref{tb.results_full_bleu,tb.results_full_chrf,tb.results_full_ter}, and align with former work \citep{gu-kong-2021-fully}. The most significant translation quality improvement can be achieved by switching from the traditional length prediction module to a \ctc-based variant, confirming the strong dependence on accurate length predictions \citep{wang-etal-2021-length}. This amounts to the largest accuracy increase that is achievable with current non-autoregressive methods but comes with a slight CPU latency increase due to the upsampling process. On top of this, \glat consistently increases translation quality further by a significant margin. Deep Supervision can further increase the quality but, for the most part, not by a significant amount. Notably, though, it rarely hurts accuracy for our experiments across all three metrics and can sometimes even achieve a significant increase on top of \ctc (\lroen on all three metrics and \lendeb on \ter). See \Cref{app:analysis} for analysis on the predicted translations.

Comparing the strongest AT and NAT system in \Cref{tab:at_vs_nat}, we can see that despite \bleu parity on WMT'14 \ldeen, the \comet scores \citep{rei-etal-2020-comet}\footnote{Obtained with \texttt{wmt20-comet-da} from version \texttt{1.1.0}} and our human evaluation results still indicate a quality gap between the two. Hence, metrics beyond \bleu are vital for rating NAT quality.

In terms of expected inference speed for non-autoregressive models, we believe that our latency numbers are more realistic than what is commonly reported in the literature. By considering additional speed-up techniques, improvements shrink to roughly $7.0\times$ on GPU and around $2.5\times$ on CPU.

\begin{table}[t]
    \centering
    \small
    \begin{tabular}{lLLLL}
    \toprule
    \multirow{3}{*}{\textbf{Metric}} & \multicolumn{4}{c}{\textbf{WMT'14}} \\
    & \multicolumn{2}{c}{\textbf{\lende}} & \multicolumn{2}{c}{\textbf{\ldeen}} \\
    & \multicolumn{1}{c}{$\Delta$} & \multicolumn{1}{c}{p-value} & \multicolumn{1}{c}{$\Delta$} & \multicolumn{1}{c}{p-value} \\
    \midrule
    \bleu $\uparrow$ & 0.9 & 0.001 & 0.0 & 0.397 \\ 
    \chrf $\uparrow$ & 0.3 & 0.009 & -0.3 & 0.002\\ 
    \ter $\downarrow$ & -1.0 & 0.001 & -1.0 & 0.001\\ 
    \comet $\uparrow$ & 0.3 & 0.000 & 0.1 & 0.000 \\ 
    Human Evaluation $\uparrow$ & 7.5 & 0.000 & 1.1 & 0.000 \\ 
    \bottomrule 
    \end{tabular}
    \caption{Score deltas (AT score $-$ NAT score) of best autoregressive vs.~non-autoregressive models on the WMT'14 \lendeb test sets with p-values of the corresponding significance tests (paired bootstrap resampling for \bleu, \chrf, \ter, \comet; paired Student's t-test for Human Acceptability Evaluation).}
    \label{tab:at_vs_nat}
\end{table}

\section{Conclusion}
In this work, we provide multiple standardized \sacrebleu metrics for non-autoregressive machine translation models, accompanied by reproducible and modular \fairseq code. We hope that our work solves comparison problems in the literature, provides more realistic speed-up expectations, and accelerates research in this area to achieve parity with autoregressive models in the future.

\newpage
\section*{Limitations}
While our work improves upon the state of translation quality and speed comparisons in the NAT literature, we acknowledge that there are many more low-level optimizations (\eg porting the Python code to \texttt{C++} and writing dedicated CUDA kernels) that could be made to improve the presented work. This would allow for an even more realistic inference speed comparison and enable direct comparability to \eg \texttt{Marian} \citep{junczys-dowmunt-etal-2018-marian} which is currently not given. There exists a concurrent work by \citet{helcl2022} that tries to address some of these shortcomings by providing a \ctc implementation in \texttt{C++} and comparing the inference speed across different batching scenarios. 

Apart from this limitation specific to our work, current non-autoregressive models still have a strong dependency on appropriate knowledge distillation from an autoregressive teacher that is able to effectively reduce data modes and limit the multimodality problem. While this already shows to be an important factor for achieving good translation quality on benchmarking datasets such as WMT, it is even more relevant for larger and/or multilingual datasets \citep{agrawal2021} that tend to be of higher complexity. Even though non-autoregressive models can lead to improved inference speed, this strong dependency still requires additional effort for tuning effective teacher architectures, conducting multiple training procedures, and generating simplified data through one or more distillation rounds. Until these problems are solved, it seems unrealistic that the average practitioner, who does not have access to a lot of compute, can effectively deploy these in production settings as general-purpose translation models. Apart from that, there is still more work needed in reducing the multimodality problem through more suitable architectures, optimization objectives, or a better data selection process.




\section*{Acknowledgements}
We would like to thank Matthias Sperber for coordinating the publication process and Yi-Hsiu Liao for providing the Shortlist implementation. Further, we'd like to thank Sarthak Garg, Hendra Setiawan, Luke Carlson, Russ Webb, and Jesse Allardice for their helpful comments on the manuscript.  Many thanks to Sachin Mehta and David Harrison for suggesting interesting ideas and the rest of Apple's Machine Translation Team for their support.


\bibliography{Bibliography/anthology,Bibliography/custom}
\bibliographystyle{acl_natbib}

\appendix

\setcounter{table}{0}
\renewcommand{\thetable}{A\arabic{table}}

\setcounter{figure}{0}
\renewcommand\thefigure{A\arabic{figure}}   

\section{Related work}
\label{app:related_work}
\looseness=-1
Many works have tried to accomplish the goal of autoregressive translation quality with a non-autoregressive model in recent years. The first paper to outline the vanilla non-autoregressive model for machine translation was \citet{Gu0XLS18}. After that, a few branches of work emerged that either focus on the \emph{multimodality problem} \citep{ma-etal-2019-flowseq, ding-etal-2020-context, qian-etal-2021-glancing, RanL0Z21, bao-etal-2021-non, song-etal-2021-alignart, huang2021nonautoregressive}, on improving the knowledge transfer from autoregressive to non-autoregressive models \citep{li-etal-2019-hint, wei-etal-2019-imitation, SunY20, ZhouGN20, GuoTXQCL20, hao-etal-2021-multi, xu-etal-2021-distilled}, on applying different training objectives to train the non-autoregressive model \citep{libovicky-helcl-2018-end, GhazvininejadKZ20, DuTJ21, shao-etal-2021-sequence}, or on incorporating language models to boost accuracy \citep{su-etal-2021-non}.

There is also a branch of work that focuses on semi-autoregressive models that try to find a balance between autoregressive and non-autoregressive models to leverage the benefits of both approaches \citep{SternSU18, lee-etal-2018-deterministic, GuWZ19, ghazvininejad-etal-2019-mask, KasaiCGG20, ChanSH0J20}. However, since they often mitigate the expected speed-up gains, we do not consider them in this work.

\section{Methods}
\label{app:methods}
In this section, we want to highlight some of the building blocks we used and explain their usage and configuration in our models. 

\subsection{Glancing Transformers}

The Glancing Transformer (GLAT, \citealp{qian-etal-2021-glancing}) tackles the \emph{multimodality problem} by encouraging word interdependence through \emph{glancing sampling}. In that, they adaptively select a number of ground truth tokens by comparing them to the model predictions $\hat{e}_1^{\hat{I}}$ and feeding the embeddings of the selected tokens to the decoder during training. Formally, the probability distribution can be described as
\begin{equation}
    p_{\bm{\theta}}(e_1^I | f_1^J) = \prod_{i=1}^I p_{\bm{\theta}}(e_i|\Phi(e_1^I, \hat{e}_1^{\hat{I}}), f_1^J),
    \label{eq.glat}
\end{equation}
where $\Phi(e_1^I, \hat{e}_1^{\hat{I}})$ represents the subset of tokens selected by the glancing sampling strategy. This sampling process can be split into two parts. First, determining the number of sampled tokens $S$ by comparing $e_1^I$ and $\hat{e}_1^{\hat{I}}$, here defined by the function $\Gamma(e_1^I, \hat{e}_1^{\hat{I}})$. Secondly, selecting $S$ tokens from the ground truth that form the final output of the glancing sampling. \citet{qian-etal-2021-glancing} deploy a random function for this purpose which we adopt and define here as the function $\Omega(e_1^I, S)$. Combining all these, the sampling strategy can be formulated as
\begin{equation}
    \Phi(e_1^I, \hat{e}_1^{\hat{I}}) = \Omega(e_1^I, \Gamma(e_1^I, \hat{e}_1^{\hat{I}})),
    \label{eq.glat_sampling}
\end{equation}
with the number of sampled tokens $S$ being determined by $\Gamma(e_1^I, \hat{e}_1^{\hat{I}}) = \lambda \cdot d(e_1^I, \hat{e}_1^{\hat{I}})$ where $d(e_1^I, \hat{e}_1^{\hat{I}})$ is a distance measure between the ground truth $e_1^I$ and the predicted tokens $\hat{e}_1^{\hat{I}}$ while $\lambda$ is a hyperparameter. For this distance measure, there are many choices that can be deployed in theory such as the Levenshtein distance \citep{Levenshtein_SPD66} or, with slight adaptions, one of the Bregman divergences \citep{BanerjeeMDG05}. For their experiments, \citet{qian-etal-2021-glancing} choose the Hamming distance \citep{Hamming1950} defined as $d(e_1^I, \hat{e}_1^{\hat{I}}) = \sum_{i=1}^I e_i \neq \hat{e}_i$ which has the useful property that it takes into account the current prediction quality of the model, causing the number of sampled tokens $S$ to be high initially and decrease over time. In their implementation, they additionally deploy a linearly decreasing schedule with a lower bound \ie $\lambda = \lambda_s - \lambda_m \cdot \frac{u}{U}$ where $u$ is the current and $U$ is the maximum number of update steps with $\lambda_s = 0.5$ and $\lambda_m = 0.2$ to further control the number of sampled tokens. We adopt their configuration.

\subsection{Connectionist Temporal Classification}

Traditional sequence-to-sequence models output target tokens autoregressively until an end-of-sentence token is encountered and are trained using the standard cross-entropy loss. In a non-autoregressive setting, though, the number of tokens in the output is not known \emph{a priori}. One solution is to train a length prediction module to determine the number of output tokens, but this approach still has drawbacks, as it doesn't properly address token repetition and the predicted length can easily be either too short or too long. The quality of the length prediction module is influenced by properties of the respective language pairs such as alignment patterns, word orders or intrinsic length ratios \citep{wang-etal-2021-length}. These fluctuations in accuracy are especially influential on the final translation quality \citep{wang-etal-2021-length}.

The Connectionist Temporal Classification (\ctc, \citealp{GravesFGS06,libovicky-helcl-2018-end}) approach tries to improve this property by setting a large enough output length $\tilde{J}$ and giving the model the option to flexibly adjust the target offsets using an additional \texttt{<blank>} token. Specifically, by marginalizing all possible alignments using dynamic programming, for all aligned sequences $a_1^{\tilde{J}}$ that can be reduced to the target $e_1^I$, the conditional probability corresponds to
\begin{equation}
    \begin{split}
    p_{\bm{\theta}}(e_1^I | f_1^J) & = \sum_{a_1^{\tilde{J}} \in \mathcal{A}(e_1^I)} p_{\bm{\theta}}(a_1^{\tilde{J}}|f_1^J) \\
    & = \sum_{a_1^{\tilde{J}} \in \mathcal{A}(e_1^I)} \prod_{j=1}^{\tilde{J}} p_{\bm{\theta}}(a_j|f_1^J)
    \end{split}
    \label{eq.ctc}
\end{equation}
where $\mathcal{A}$ is a one-to-many map that produces all possible alignments $a_1^{\tilde{J}}$ and $\mathcal{A}^{-1}$ is a many-to-one map that collapses an alignment to recover the target sequence by first removing repeated and then \texttt{<blank>} tokens \eg 
\begin{equation}
\mathcal{A}^{-1}(\underbrace{\bigstrut[b]a-abb}_{a_1^5}) = \mathcal{A}^{-1}(\underbrace{\bigstrut[b]aa-ab}_{a_1^5}) = \underbrace{\bigstrut[b]aab}_{e_1^3}
\label{eq.ctc_example}
\end{equation}
with $-$ representing the \texttt{<blank>} token for brevity. This approach requires setting an alignment length $\tilde{J}$ that is determined by a scaling factor $s$ commonly chosen to be $2$ -- $3$ times the source length $J$, i.e.~$\tilde{J} = J \cdot s$. In the example of \Cref{eq.ctc_example}, the alignment length is $\tilde{J} = 5$.

\paragraph{Combining with \glat}
To effectively combine \ctc with \glat, an alignment for the target tokens is required to enable the glancing sampling correspondence with the decoder input tokens. For that, \citet{gu-kong-2021-fully} suggest using the Viterbi aligned tokens  
\begin{equation}
    \hat{a}_1^{\tilde{J}}(f_1^J) = \argmax_{a_1^{\tilde{J}} \in \mathcal{A}(e_1^I)}\ p_{\bm{\theta}}(a_1^{\tilde{J}}|f_1^J)
\end{equation}
which we adopt in our implementation and obtain leveraging the \texttt{best\_alignment} method from the Imputer\footnote{\url{https://github.com/rosinality/imputer-pytorch}} \citep{ChanSH0J20}. Similar to previous works, we use greedy decoding instead of beam search for all our experiments to keep a strict non-autoregressive property.

\subsection{Deep Supervision}
\label{sec.ds}
The Deeply Supervised Layer-wise Prediction-aware Transformer (DSLP, \citealp{huang2021nonautoregressive}) introduces three strategies to address the \emph{multimodality problem}. Namely, Deep Supervision, Layer-Wise Prediction-Awareness, and Mixed Training. In a nutshell, Deep Supervision refers to predicting the output tokens at every decoder layer, computing the layer-wise losses, and aggregating them to form the overall loss used for optimizing the model. Formally, we can maximize the log-likelihood
\begin{equation}
    \mathcal{L} = \frac{1}{L_D} \sum_{l=1}^{L_D} \sum_{i=1}^I \log p_{\bm{\theta}}^{(l)}(e_i^{(l)}| \bm{h}^{(l)}_i),
    \label{eq.deep_supervision}
\end{equation}
where $\bm{h}^{(l)}_i$ is the hidden state of decoder layer $l$ corresponding to position $i$ and $L_D$ is the overall number of decoder layers. Intuitively, this approach loses its value when moving to a deep encoder and shallow decoder \citep{Kasai0PCS21} with $L_D = 1$. 

While Deep Supervision keeps the model parameter count constant and only increases the used memory, the other two proposed techniques increase the number of trained parameters. For this work, we only adopt the Deep Supervision part due to the aforementioned drawback.

\section{Ablations}


\subsection{Influence of dropout}
\label{app:dropout}
\begin{figure*}
\centering

\begin{subfigure}{0.48\textwidth}
    \includegraphics[width=\textwidth]{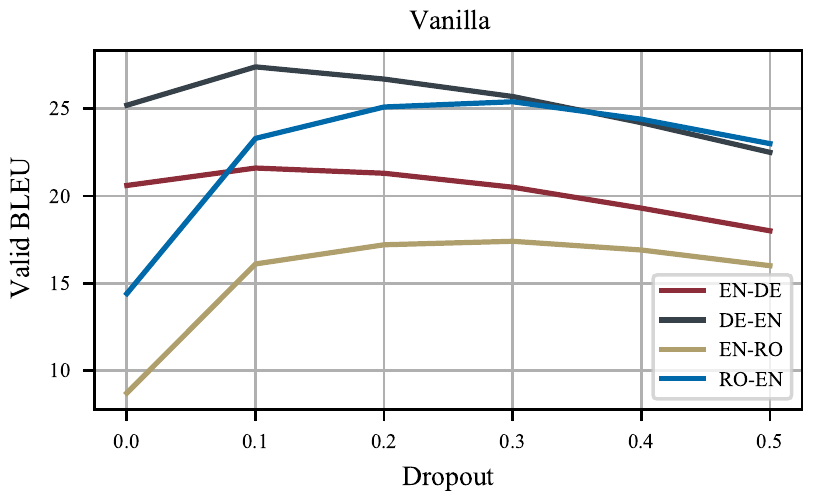}
\end{subfigure}
\hfill
\begin{subfigure}{0.48\textwidth}
    \includegraphics[width=\textwidth]{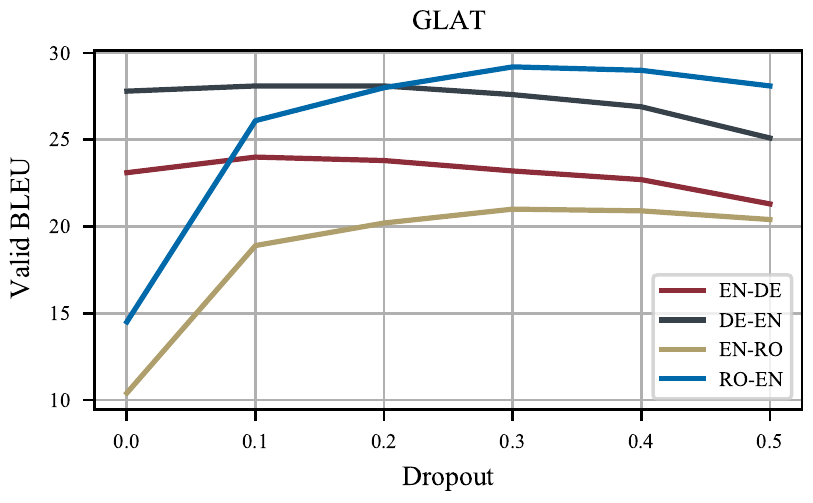}
\end{subfigure}

\begin{subfigure}{0.48\textwidth}
    \includegraphics[width=\textwidth]{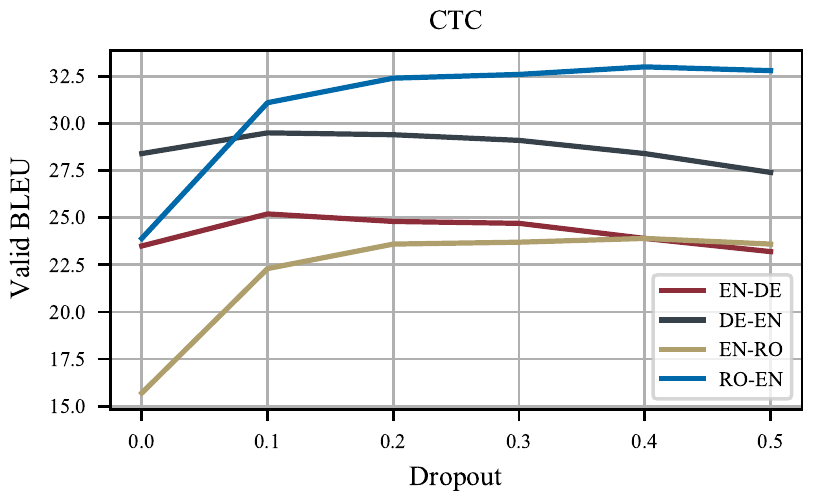}
\end{subfigure}
\hfill
\begin{subfigure}{0.48\textwidth}
    \includegraphics[width=\textwidth]{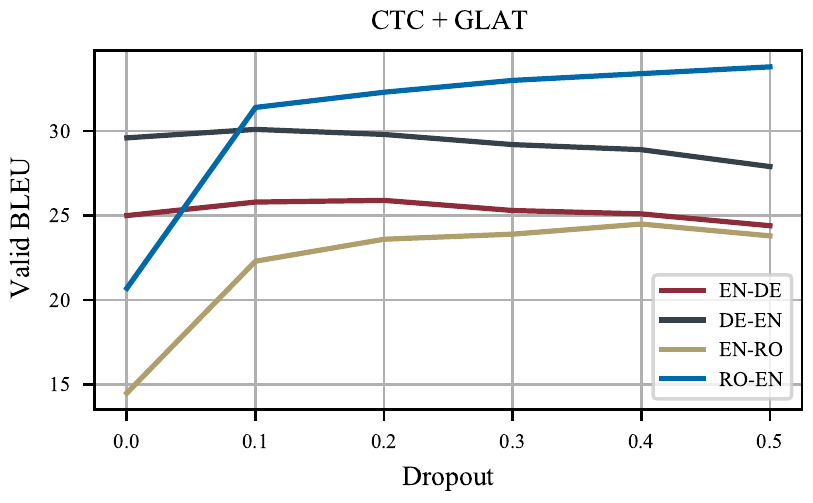}
\end{subfigure}
\caption{Influence of dropout for different models and language pairs on the respective validation set.}
 \label{fig.drop_swp}
\end{figure*}

\Cref{fig.drop_swp} shows the \bleu score plotted against the dropout rate for all translation directions and multiple architectures. In general, our results agree with the choices of prior work with $0.1$ performing best for \lendeb, and $0.3$ for \lenrob. For the \ctc-based methods, higher dropouts seem to give slightly better results.

\subsection{Optimization choices}
\label{app:adam_eps}
We conduct experiments to assess the value of more optimization choices in \Cref{tb.flag_ablation}. From these, it seems that most of them seem to have a positive impact when being deployed for non-\ctc variants while they don't seem to benefit \ctc-based variants.

\begin{table}[ht]
\centering
\small
\tabcolsep 4pt
\begin{tabular}{lcc}
\toprule
\multirow{2}{*}{\textbf{Models}} & \textbf{WMT'13} & \textbf{WMT'14} \\
 & \textbf{\lende} & \textbf{\lende} \\
 \midrule
 \textbf{Vanilla NAT}  & \multirow{2}{*}{21.1} & \multirow{2}{*}{20.0} \\ 
 \citep{Gu0XLS18} & & \\[0.1cm]
 \quad\quad + GELU & 21.4 \chu{0.3} & 20.0 \che{0.0} \\ 
 \quad\quad + \bert init & 21.3 \chu{0.2} & 20.1 \chu{0.1} \\ 
 \quad\quad + Predict length offset & 21.4 \chu{0.3} & 20.0 \che{0.0} \\ 
 \quad\quad + \adam $\epsilon=1e-6$ & 21.2 \chu{0.1} & 20.1 \chu{0.1} \\ 
 \midrule
 \textbf{\ctc} & \multirow{2}{*}{25.0} & \multirow{2}{*}{25.6} \\  
 \citep{saharia-etal-2020-non} & & \\[0.1cm]
 \quad\quad + GELU & 25.0 \che{0.0} & 25.5 \chd{0.1} \\ 
 \quad\quad + \bert init & 24.8 \chd{0.2} & 25.6 \che{0.0} \\ 
 \quad\quad + \adam $\epsilon=1e-6$ & 25.2 \chu{0.2} & 25.7 \chu{0.1} \\ 
 \bottomrule
\end{tabular}
\caption{Overview of the quality implications when adding different optimization techniques. Distilled training data and hyperparameters stay consistent and results are reported for the evaluation sets of WMT'13 and WMT'14 for \lende.}
\label{tb.flag_ablation}
\end{table}

\Cref{tb.flag_ablation} suggests that choosing a slightly higher \adam $\epsilon$ might have a positive effect for training non-autoregressive models. To this end, we ran a grid search in the range $[1\mathrm{e}{-9}, 1\mathrm{e}{-1}]$ with a delta factor between samples of $\Delta = 10^{-1}$ for \ctc + \glat, one of our top-performing models, to verify the observed gains. The results of this experiment can be seen in \Cref{tb.eps_ablation}.
 
\begin{table}[ht]
\centering
\small
\tabcolsep 4pt
\begin{tabular}{lcc}
\toprule
\multirow{2}{*}{\textbf{Models}} & \textbf{WMT'13} & \textbf{WMT'14} \\
 & \textbf{\lende} & \textbf{\lende} \\
 \midrule
 \textbf{\ctc + \glat} & \multirow{2}{*}{25.8} & \multirow{2}{*}{25.9} \\  
 \citep{qian-etal-2021-glancing} & & \\[0.1cm] 
 \quad\quad + \emph{\adam $\epsilon=1e-9$} & 25.9 \chu{0.1} & 25.9 \che{0.0} \\
 \quad\quad + \emph{\adam $\epsilon=1e-7$} & 25.8 \che{0.0} & 26.0 \chu{0.1} \\ 
 \quad\quad + \emph{\adam $\epsilon=1e-6$} & 25.7 \chd{0.1} & 25.9 \che{0.0} \\
 \quad\quad + \emph{\adam $\epsilon=1e-5$} & 25.7 \chd{0.1} & 25.6 \chd{0.3} \\ 
 \quad\quad + \emph{\adam $\epsilon=1e-4$} & 25.3 \chd{0.5} & 25.1 \chd{0.8} \\ 
 \quad\quad + \emph{\adam $\epsilon=1e-3$} & 22.0 \chd{3.8} & 21.2 \chd{4.7} \\ 
 \quad\quad + \emph{\adam $\epsilon=1e-2$} & \emph{diverges} & \emph{diverges} \\ 
 \quad\quad + \emph{\adam $\epsilon=1e-1$} & \emph{diverges} & \emph{diverges} \\ 
 \bottomrule
\end{tabular}
\caption{Overview of the quality implications when changing the \adam $\epsilon$ for \ctc + \glat. The baseline run uses the default of $\epsilon = 1\mathrm{e}{-8}$. }
\label{tb.eps_ablation}
\end{table}

From our results we conclude that the observed gains are neither consistent nor statistically significant as they don't transfer to the \ctc + \glat model without further tuning of the remaining hyperparamaters such as the learning rate $\eta$ or the \adam momentum terms $\beta_1$ and $\beta_2$. To simplify hyperparameter transfer to evolving production datasets and randomness in the training procedure without any additional tuning, we keep the default of $\epsilon = 1e-8$ (similar to \citealp{SchmidtSH21}) for our models that use a combination of methods such as \ctc + \glat or \ctc + \ds and advise the community to avoid this type of tuning for comparability. While it may seem appealing at first, it can distort expected translation quality gains for novel methods as commonly this sort of tuning is only conducted for the introduced method at hand but \emph{not}, or only marginally, for baselines or competing algorithms. Any gain that is achieved through such practices should \emph{not} be accredited to the introduced method as other algorithms are most likely able to achieve similar gains through an equal amount of hyperparameter tuning.

\section{Result details}
\label{app:full_results}

\subsection{Dataset}
\label{app:dataset_details}
We preprocessed the datasets using Moses scripts \citep{koehn-etal-2007-moses} to tokenize all datasets, clean the data (\texttt{clean-corpus-n.pl}), and normalize punctuation (\texttt{normalize-punctuation.perl}). No True-Casing is used. For \lroen we normalize Romanian and remove diacritics using WMT'16 provided scripts\footnote{\url{https://github.com/rsennrich/wmt16-scripts/tree/master/preprocess}}. \lenro keeps diacritics for producing accurate translations which explains the gap between our numbers and those claimed in previous works \citep{gu-kong-2021-fully,qian-etal-2021-glancing,Gu0XLS18}, who besides computing \bleu on tokenized text, compared on diacritic-free text.

For ablations, which are only conducted on WMT'14 \lende distilled training data, we additionally report accuracy on the WMT'13 \lende evaluation set, serving as validation performance. We use a shared source and target vocabulary for all language directions, obtained using BPE with $40$k merge operations where the distillation process dropped around $400-800$ tokens depending on the respective target language. 

\subsection{Signatures: \sacrebleu}
\label{app:signatures}
We use \sacrebleu version $2.0.0$. The evaluation signatures for \bleu \citep{papineni-etal-2002-bleu} are \texttt{nrefs:1} \texttt{|} \texttt{case:mixed} \texttt{|} \texttt{eff:no} \texttt{|} \texttt{tok:13a} \texttt{|} \texttt{smooth:exp}. Additionally we also report \chrf \citep{popovic-2017-chrf} with \texttt{nrefs:1} \texttt{|} \texttt{case:mixed} \texttt{|} \texttt{eff:yes} \texttt{|} \texttt{nc:6} \texttt{|} \texttt{nw:2} \texttt{|} \texttt{space:no} and \ter \citep{snover-etal-2006-study} with \texttt{nrefs:1} \texttt{|} \texttt{case:mixed} \texttt{|} \texttt{tok:tercom} \texttt{|} \texttt{norm:no} \texttt{|} \texttt{punct:yes} \texttt{|} \texttt{asian:no} for the main results.

\subsection{Paired boostrap resampling}
\label{sec:stat_tests}
For the boostrap resampling used to identify \insig results in the main tables, we always use the root level of the sub-blocks (separated by dashed lines) as base system within that specific sub-block. For models that deploy multiple additional methods, \eg \ctc + \glat + \ds, we always use the previous row as the base system, \eg \ctc + \glat in this case. This procedure is especially relevant for the significance tests on the autoregressive baselines that use knowledge distillation, average attention, and shortlists and helps to show the significance of the individual methods on the translation quality. While we want significant results for the NAT category, it is beneficial to have \insig results for the methods that are deployed to speed-up CPU inference (\ie average attention and shortlists) as this shows that the respective model is not statistically different while offering faster CPU inference. Finally, for the root levels of each sub-block we use the first sub-block of each category as the base system \ie both \ctc and \ctc (11-2) are compared to Vanilla-NAT while Transformer \emph{base} (11-2) is compared to Transformer \emph{base}. The null hypothesis is always that the reference system and the baseline translations are essentially generated by the same underlying process. If they are marked with $\nosig$, the null hypothesis could not be rejected.

\paragraph{Additional Metrics}
Both \Cref{tb.results_full_chrf,tb.results_full_ter}, which report \chrf and \ter, support our conclusions drawn in \Cref{sec.results} of the main paper, statistical significance sometimes deviates marginally.

\begin{table*}[!ht]
\setlength\tabcolsep{4pt}
\centering
\scriptsize
\begin{tabular}{llLLLLLrS[table-format=3.1]@{${}\pm{}$}S[table-format=1.2]rS[table-format=4.1]@{${}\pm{}$}S[table-format=2.1]}
\toprule
 \multicolumn{1}{l}{\multirow{2}{*}{\textbf{Models}}} & \multicolumn{1}{c}{\multirow{2}{*}{$|\bm{\theta}|$}} & 
 \multicolumn{1}{c}{\textbf{WMT'13}} & \multicolumn{2}{c}{\textbf{WMT'14}} & \multicolumn{2}{c}{\textbf{WMT'16}} &
 \multicolumn{1}{c}{\multirow{1}{*}{\textbf{Speed}}} & \multicolumn{2}{c}{\multirow{1}{*}{\textbf{Latency GPU}}} & \multicolumn{1}{c}{\multirow{1}{*}{\textbf{Speed}}} & \multicolumn{2}{c}{\multirow{1}{*}{\textbf{Latency CPU}}} \\
 \multicolumn{1}{c}{} & & \multicolumn{1}{c}{\textbf{\lende}} & \multicolumn{1}{c}{\textbf{\lende}} & \multicolumn{1}{c}{\textbf{\ldeen}} & \multicolumn{1}{c}{\textbf{\lenro}} & \multicolumn{1}{c}{\textbf{\lroen}} & \multicolumn{1}{c}{(GPU)} & \multicolumn{2}{c}{(\si{\ms}~/~sentence)} &
 \multicolumn{1}{c}{(CPU)} & 
 \multicolumn{2}{c}{(\si{\ms}~/~sentence)}\\
\midrule
\textbf{Autoregressive} & & & & & & & & \multicolumn{1}{c}{} & & & \multicolumn{1}{c}{} & \\
\midrule

 Transformer \textit{base} & 66M & 51.7 & 54.1 & 55.4 & 50.4 & 57.0 & 0.5$\times$ & 165.8 & 1.3  & 0.3$\times$ & 657.2 & 11.4 \\
 \quad\quad + Beam $5$ (teacher) & 66M & 52.3 & 54.8 & 56.3 & 50.9 & 57.5 & 0.4$\times$ & 212.8 & 8.9  & 0.1$\times$ & 1272.4 & 38.5 \\
 \quad\quad + KD & 66M & 52.2 & 54.8 & 55.6 & 50.7 & 57.1\nosig & 0.5$\times$ & 173.4 & 1.0  & 0.3$\times$ & 646.3 & 8.9 \\
\cdashlinelr{1-13}
 Transformer \textit{base} (11-2) & 63M $\downarrow$ & 51.6\nosig & 54.1\nosig & 55.0 & 50.5\nosig & 57.0\nosig & 0.9$\times$ & 89.1 & 3.2  & 0.4$\times$ & 424.2 & 15.0 \\
 \quad\quad + KD & 63M $\downarrow$ & 52.2 & 54.9 & 56.0 & 50.6\nosig & 57.1\nosig & 0.9$\times$ & 88.1 & 1.5  & 0.5$\times$ & 408.0 & 3.2 \\
 \quad\quad + KD + AA & 62M $\downarrow$ & 52.2\nosig & 54.8\nosig & 55.7 & 50.7\nosig & 57.0\nosig & 1.0$\times$ & 79.4 & 2.2  & 0.5$\times$ & 396.9 & 5.4 \\
 \quad\quad + KD + AA + SL & 62M $\downarrow$ & 52.2\nosig & 54.8\nosig & 55.7 & 50.7\nosig & 57.0 & 1.0$\times$ & 82.7 & 7.5  & 1.0$\times$ & 188.3 & 2.2 \\

\midrule
\textbf{Non-Autoregressive} & & & & & & & & \multicolumn{1}{c}{} & & & \multicolumn{1}{c}{} & \\
\midrule

 Vanilla-NAT & 66M & 48.5 & 50.2 & 52.5 & 46.6 & 52.8 & 6.1$\times$ & 13.1 & 0.3  & 2.1$\times$ & 90.1 & 0.4 \\
 \quad\quad + \glat & 66M & 50.5 & 52.7 & 54.4 & 48.4 & 54.5 & 6.2$\times$ & 12.8 & 0.4  & 2.1$\times$ & 90.4 & 0.6 \\
\cdashlinelr{1-13}
 \ctc & 66M & 51.5 & 53.8 & 55.2 & 50.3 & 57.1 & 6.4$\times$ & 12.4 & 0.4  & 1.5$\times$ & 122.9 & 1.7 \\
 \quad\quad + \ds & 66M & 51.5\nosig & 53.9\nosig & 55.3\nosig & 50.3\nosig & 57.3 & 6.3$\times$ & 12.7 & 0.6  & 1.6$\times$ & 121.4 & 0.5 \\
 \quad\quad + \glat  & 66M & 52.0 & 54.4 & 55.7 & 50.6 & 57.3 & 6.3$\times$ & 12.7 & 0.3  & 1.6$\times$ & 121.2 & 1.2 \\
 \quad\quad + \glat + \ds & 66M & 52.1 & 54.6 & 55.9 & 50.7\nosig & 57.3\nosig & 6.2$\times$ & 12.8 & 0.3  & 1.5$\times$ & 121.8 & 0.4 \\
\cdashlinelr{1-13}
 \ctc (11-2) & 63M $\downarrow$ & 51.6 & 54.0 & 55.4 & 50.6 & 57.4 & 6.9$\times$ & 11.5 & 0.2  & 1.6$\times$ & 114.6 & 1.3 \\
 \quad\quad + \glat & 63M $\downarrow$ & 52.0 & 54.5 & 56.0 & 51.0 & 57.6 & 6.8$\times$ & 11.7 & 0.4  & 1.6$\times$ & 114.6 & 1.2 \\
 \quad\quad + \glat + SL & 63M $\downarrow$ & 52.0\nosig & 54.5\nosig & 56.0 & 51.0\nosig & 57.6\nosig & 6.7$\times$ & 11.9 & 0.2  & 2.6$\times$ & 72.5 & 0.8 \\

\bottomrule
\end{tabular}
\caption{Main results using \chrf on four different translation directions. Latencies are measured on the WMT'14 \lende test set as the average processing time per sentence using batch size $1$. All AT models require $I$ decoding iterations at inference time while the NAT models only require $1$. WMT'13 \lende serves as validation performance. No re-scoring is used for all results. Statistically \insig results using paired bootstrap resampling are marked with $\nosig$ for $p \geq 0.05$, see \Cref{sec:stat_tests} for more details. \lenro keeps diacritics and $\downarrow$ indicates less number of parameters than the baseline.}
\label{tb.results_full_chrf}
\end{table*}

\begin{table*}[!ht]
\setlength\tabcolsep{4pt}
\centering
\scriptsize
\begin{tabular}{llLLLLLrS[table-format=3.1]@{${}\pm{}$}S[table-format=1.2]rS[table-format=4.1]@{${}\pm{}$}S[table-format=2.1]}
\toprule
 \multicolumn{1}{l}{\multirow{2}{*}{\textbf{Models}}} & \multicolumn{1}{c}{\multirow{2}{*}{$|\bm{\theta}|$}} & 
 \multicolumn{1}{c}{\textbf{WMT'13}} & \multicolumn{2}{c}{\textbf{WMT'14}} & \multicolumn{2}{c}{\textbf{WMT'16}} &
 \multicolumn{1}{c}{\multirow{1}{*}{\textbf{Speed}}} & \multicolumn{2}{c}{\multirow{1}{*}{\textbf{Latency GPU}}} & \multicolumn{1}{c}{\multirow{1}{*}{\textbf{Speed}}} & \multicolumn{2}{c}{\multirow{1}{*}{\textbf{Latency CPU}}} \\
 \multicolumn{1}{c}{} & & \multicolumn{1}{c}{\textbf{\lende}} & \multicolumn{1}{c}{\textbf{\lende}} & \multicolumn{1}{c}{\textbf{\ldeen}} & \multicolumn{1}{c}{\textbf{\lenro}} & \multicolumn{1}{c}{\textbf{\lroen}} & \multicolumn{1}{c}{(GPU)} & \multicolumn{2}{c}{(\si{\ms}~/~sentence)} &
 \multicolumn{1}{c}{(CPU)} & 
 \multicolumn{2}{c}{(\si{\ms}~/~sentence)}\\
\midrule
\textbf{Autoregressive} & & & & & & & & \multicolumn{1}{c}{} & & & \multicolumn{1}{c}{} & \\
\midrule

 Transformer \textit{base} & 66M & 64.1 & 63.0 & 56.9 & 64.4 & 55.6 & 0.5$\times$ & 165.8 & 1.3  & 0.3$\times$ & 657.2 & 11.4 \\
 \quad\quad + Beam $5$ (teacher) & 66M & 63.7 & 62.4 & 56.0 & 63.8 & 55.1 & 0.4$\times$ & 212.8 & 8.9  & 0.1$\times$ & 1272.4 & 38.5 \\
 \quad\quad + KD & 66M & 64.0\nosig & 62.9\nosig & 56.3 & 64.0 & 55.5\nosig & 0.5$\times$ & 173.4 & 1.0  & 0.3$\times$ & 646.3 & 8.9 \\
\cdashlinelr{1-13}
 Transformer \textit{base} (11-2) & 63M $\downarrow$ & 63.8\nosig & 63.1\nosig & 57.0\nosig & 63.9 & 55.4\nosig & 0.9$\times$ & 89.1 & 3.2  & 0.4$\times$ & 424.2 & 15.0 \\
 \quad\quad + KD & 63M $\downarrow$ & 63.7\nosig & 62.4 & 55.6 & 63.7\nosig & 55.3\nosig & 0.9$\times$ & 88.1 & 1.5  & 0.5$\times$ & 408.0 & 3.2 \\
 \quad\quad + KD + AA & 62M $\downarrow$ & 63.8\nosig & 62.4\nosig & 55.4\nosig & 63.6\nosig & 55.7\nosig & 1.0$\times$ & 79.4 & 2.2  & 0.5$\times$ & 396.9 & 5.4 \\
 \quad\quad + KD + AA + SL & 62M $\downarrow$ & 63.8\nosig & 62.4\nosig & 55.4 & 63.6\nosig & 55.7 & 1.0$\times$ & 82.7 & 7.5  & 1.0$\times$ & 188.3 & 2.2 \\

\midrule
\textbf{Non-Autoregressive} & & & & & & & & \multicolumn{1}{c}{} & & & \multicolumn{1}{c}{} & \\
\midrule

 Vanilla-NAT & 66M & 67.9 & 68.1 & 60.1 & 68.0 & 59.4 & 6.1$\times$ & 13.1 & 0.3  & 2.1$\times$ & 90.1 & 0.4 \\
 \quad\quad + \glat & 66M & 65.5 & 65.1 & 58.0 & 66.2 & 57.8 & 6.2$\times$ & 12.8 & 0.4  & 2.1$\times$ & 90.4 & 0.6 \\
\cdashlinelr{1-13}
 \ctc & 66M & 64.5 & 63.8 & 56.6 & 64.5 & 55.6 & 6.4$\times$ & 12.4 & 0.4  & 1.5$\times$ & 122.9 & 1.7 \\
 \quad\quad + \ds & 66M & 64.2 & 63.4 & 56.1 & 64.5\nosig & 55.3 & 6.3$\times$ & 12.7 & 0.6  & 1.6$\times$ & 121.4 & 0.5 \\
 \quad\quad + \glat  & 66M & 64.1 & 63.5\nosig & 56.0 & 64.0 & 55.1 & 6.3$\times$ & 12.7 & 0.3  & 1.6$\times$ & 121.2 & 1.2 \\
 \quad\quad + \glat + \ds & 66M & 64.1\nosig & 63.3\nosig & 55.9\nosig & 64.2\nosig & 55.4\nosig & 6.2$\times$ & 12.8 & 0.3  & 1.5$\times$ & 121.8 & 0.4 \\
\cdashlinelr{1-13}
 \ctc (11-2) & 63M $\downarrow$ & 64.0 & 63.5 & 56.3 & 64.1 & 54.9 & 6.9$\times$ & 11.5 & 0.2  & 1.6$\times$ & 114.6 & 1.3 \\
 \quad\quad + \glat & 63M $\downarrow$ & 64.4 & 63.4\nosig & 56.4\nosig & 63.7 & 55.1\nosig & 6.8$\times$ & 11.7 & 0.4  & 1.6$\times$ & 114.6 & 1.2 \\
 \quad\quad + \glat + SL & 63M $\downarrow$ & 64.4\nosig & 63.4\nosig & 56.4\nosig & 63.7\nosig & 55.1\nosig & 6.7$\times$ & 11.9 & 0.2  & 2.6$\times$ & 72.5 & 0.8 \\

\bottomrule
\end{tabular}
\caption{Main results using \ter on four different translation directions. Latencies are measured on the WMT'14 \lende test set as the average processing time per sentence using batch size $1$. All AT models require $I$ decoding iterations at inference time while the NAT models only require $1$. WMT'13 \lende serves as validation performance. No re-scoring is used for all results. Statistically \insig results using paired bootstrap resampling are marked with $\nosig$ for $p \geq 0.05$, see \Cref{sec:stat_tests} for more details. \lenro keeps diacritics and $\downarrow$ indicates less number of parameters than the baseline.}
\label{tb.results_full_ter}
\end{table*}

\section{Analysis}
\label{app:analysis}
\begin{figure}[!ht]
    \centering
    \includegraphics[width=\columnwidth]{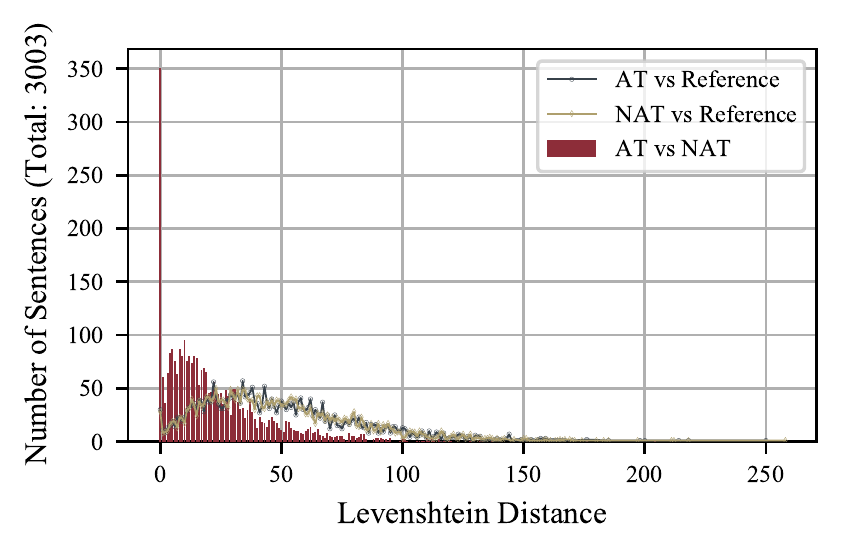}
    \caption{Levenshtein Distance between the hypotheses generated from the autoregressive Transformer \emph{base} (11-2) + KD and \ctc + \glat (11-2) on the WMT'14 \ldeen test set. Here, $351$ hypotheses match exactly. Individual distances of the hypotheses against the ground truth are plotted for reference.}
    \label{fig:levenshtein}
\end{figure}

\begin{figure}[!ht]
    \centering
    \includegraphics[width=\columnwidth]{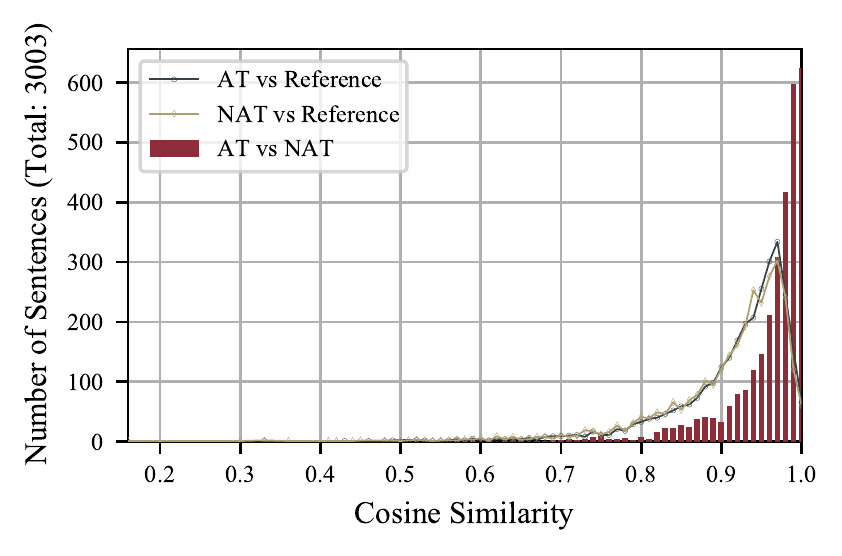}
    \caption{Cosine similarity for BERT hypothesis embeddings (mean final hidden state) generated from the autoregressive Transformer \emph{base} (11-2) + KD and \ctc + \glat (11-2) on the WMT'14 \ldeen test set.}
    \label{fig:cosine_sim}
\end{figure}

\paragraph{Hypotheses characteristics}
Comparing one of the best performing NAT models, \ctc + \glat (11-2) to its autoregressive counterpart Transformer \emph{base} (11-2), we try to understand the nature of occurring errors and their similarity. For that, we compare the Levenshtein distance between each models' hypotheses and the reference as well as their hypotheses with each other in \Cref{fig:levenshtein} on the WMT'14 \ldeen test set. Both AT and NAT follow a similar distribution of distance when compared to the reference. However, when comparing their hypothesis to each other, only $351/3003$ hypotheses match exactly and often we can observe Levenshtein distances between $0-50$. If we compare their sentence embeddings using BERT from \texttt{sentence-transformers} \citep{reimers-gurevych-2019-sentence} in \Cref{fig:cosine_sim}, we observe a similar trend when comparing them to the reference, although the autoregressive model is able to achieve slightly higher cosine similarity. Looking at samples with low cosine similarity between AT and NAT, we observe NAT problems that have been pointed out by prior work including repeated tokens and non-coherent sentences.

\begin{figure}[!ht]
    \centering
    \includegraphics[width=\columnwidth]{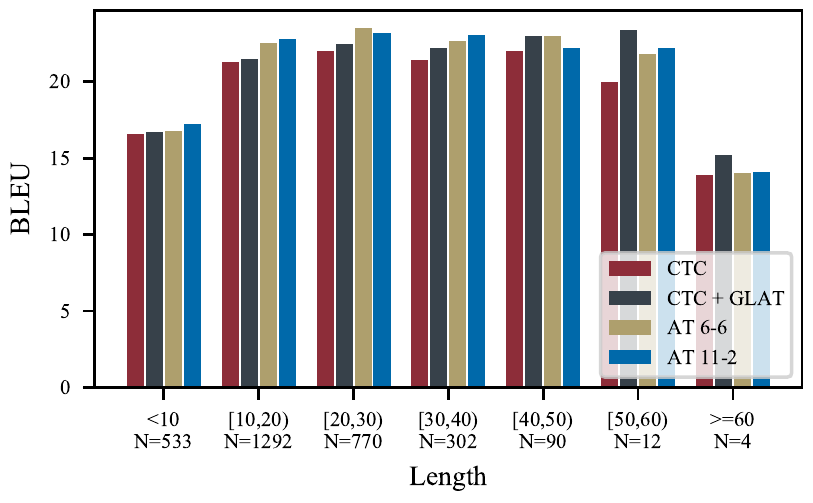}
    \caption{Effect of sentence length on \bleu for multiple architectures on WMT'14 \lende test set. The height of each bucket reflects the corpus-level \bleu for that bucket, computed using \mtcompare \citep{neubig-etal-2019-compare}. $N$ indicates the number of sentences in each bucket.}
    \label{fig:bleu_vs_length}
\end{figure}

Looking at the effect of sentence length on \bleu in \Cref{fig:bleu_vs_length}, we observe that for sequence lengths $<40$ the autoregressive models outperform the non-autoregressive models while for sequence lengths $>40$ we can see \ctc + \glat outperforming. We attribute this to the significantly smaller sample size for longer sentences.

\paragraph{Human Evaluation}
In addition to comparing the best AT model against the best NAT model based on automatic metrics such as \bleu, \chrf, \ter and \comet, we conducted a human evaluation.
To determine the human preference between AT and NAT model output, we showed professional translators both translations together with the source sentence in a side-by-side acceptability evaluation.
We asked the translators to provide an acceptability score between 0 (nonsense) and 100 (perfect translation) for each translation.
As a guideline, which was provided to the translators, a score higher than 66 means the translation retains core meaning with minor mistakes. Scores lower or equal to 66 should be assigned to translations which preserve only some meaning with major mistakes.
The translators further had the option to provide comments about why they assigned a specific score and/or preferred a certain translation.
Each of the 3003 sentences in the WMT'14 test sets was annotated once.

\end{document}